\documentclass[11pt]{article}
\usepackage{acl}
\usepackage{times}
\usepackage{latexsym}
\usepackage[T1]{fontenc}
\usepackage[utf8]{inputenc}
\usepackage{microtype}
\usepackage{booktabs}
\usepackage{multirow}
\usepackage{graphicx}
\graphicspath{{figures/}}
\usepackage{amsmath}
\usepackage{amssymb}
\usepackage{xcolor}
\usepackage{array}
\usepackage{hyperref}
\usepackage{url}
\usepackage{enumitem}
\usepackage{pifont}  
\usepackage[most]{tcolorbox}
\usepackage[framemethod=tikz]{mdframed}
\usepackage{tikz,pgfplots}

\newcommand{\sipw}{\text{SIPW}}    
\newcommand{\gain}{\Delta\mathrm{acc}}  
\newcommand{\feat}{f}              
\newcommand{\tasktype}{\tau}       

\usepackage{algorithm,algorithmic}
\newcommand{\circleone}[1]{%
    \resizebox{!}{0.8em}{%
        \tikz[baseline=(char.base)]{
            \node[shape=circle, fill=black, inner sep=0.8pt, text=white] (char) {#1};
        }%
    }%
}

\newcommand{\circletwo}[1]{%
    \resizebox{!}{0.8em}{%
        \tikz[baseline=(char.base)]{
            \node[shape=circle, fill=black, inner sep=0.8pt, text=white] (char) {#1};
        }%
    }%
}

\newcommand{\circlethree}[1]{%
    \resizebox{!}{0.8em}{%
        \tikz[baseline=(char.base)]{
            \node[shape=circle, fill=black, inner sep=0.8pt, text=white] (char) {#1};
        }%
    }%
}

\newcommand{\circlefour}[1]{%
    \resizebox{!}{0.8em}{%
        \tikz[baseline=(char.base)]{
            \node[shape=circle, fill=black, inner sep=0.8pt, text=white] (char) {#1};
        }%
    }%
}

\newcommand{\circlefive}[1]{%
    \resizebox{!}{0.8em}{%
        \tikz[baseline=(char.base)]{
            \node[shape=circle, fill=black, inner sep=0.8pt, text=white] (char) {#1};
        }%
    }%
}

\title{Why Prompt Optimization Works, and Why It Sometimes Doesn’t  \\
 — A Causal-Inspired Edit-Level Analysis}

\author{
  Shuzhi Gong\textsuperscript{\rm 1}, Hechuan Wen\textsuperscript{\rm 2}\thanks{     
    $\,$ Corresponding author.}, 
  \\
  \textsuperscript{\rm 1} The University of Melbourne,  Melbourne, Victoria, Australia\\
  \textsuperscript{\rm 2} The University of Queensland, Brisbane, Queensland, Australia \\
\texttt{shuzhi@unimelb.edu.au} \\ 
    \texttt{h.wen@uq.edu.au}}

\begin{document}
\maketitle

\begin{abstract}

Automated prompt optimization methods (e.g., DSpy, TextGrad) can substantially improve the performance of large language model (LLM), however, their generalization ability across different tasks remains underperformed. In practice, the superiority of the optimized prompt on one benchmark often fails to transfer to another, and this limitation persists even when switching across different LLM backbones. To investigate the underexplored sources of heterogeneity in prompt performance, we conduct a causal inference-inspired observational analysis of optimized prompts across a diverse set of optimization frameworks, LLM backbones, and NLP benchmarks. To achieve the goal, we build upon the propensity-adjusted associational analysis together with multiple complementary representations of prompt edits, where the consistent task-conditioned edits patterns are identified. We find that complexity-increasing and meta-instructional edits are negatively associated with mathematical and multi-hop reasoning performance, whereas step-by-step and meta-cognitive edits improve logical and sequential reasoning tasks. These effects are robust across cognitive-load annotations, surface-level text features, and edit-motif analyses, and can generalize across optimization frameworks. Overall, these results indicate that prompt optimization failures arise from systematic interactions between edit families and task characteristics rather than random optimization artifacts, providing feature-level characterization of optimizer behavior and motivating future task-conditioned optimizer design.

\end{abstract}

\section{Introduction}

Prompt optimization has emerged as an increasingly practical alternative to parameter-efficient fine-tuning for large language models (LLMs)~\cite{khattab2024dspy,opsahlong2024miprov2,yuksekgonul2024textgrad,gepa}. Instead of updating model weights, recent frameworks such as TextGrad~\cite{yuksekgonul2024textgrad}, GEPA~\cite{gepa} automatically search the prompt space to improve downstream task performance. These approaches have become particularly attractive for modern LLM systems because they are lightweight, modular, and directly compatible with LLM and agentic applications~\cite{opsahl2024optimizing,vldb-automatic,susnjak2026reproducible} and retrieval-augmented workflows~\cite{camara2026self,gong2026multi}.

Despite their empirical success, prompt optimizers often exhibit unstable behavior across tasks and model backbones~\cite{zhang2026coinflip,fu2026textreg,singhal2026prefpo}. In practice, prompt revisions that improve one benchmark frequently fail to transfer to another, and optimizers that perform well on logical or sequential reasoning tasks may substantially degrade performance on mathematical or multi-hop reasoning benchmarks. Importantly, this pattern appears consistently in our experiments across multiple optimization frameworks and across diverse LLM backbones, including GPT-5.2, GPT-4o~\cite{achiam2023gpt}, Qwen3-32B~\cite{yang2025qwen3}, and Deepseek~\cite{liu2024deepseek} models. Such instability raises a fundamental question: \textit{what kinds of prompt edits are modern optimizers actually learning to apply}, and \textit{why do some edit patterns help certain task types while harming others}?

Existing work has largely studied prompt optimization at the aggregate benchmark level~\cite{wan2024teach}. Prior studies analyze optimizer success rates~\cite{pryzant2023automatic}, optimization dynamics~\cite{yuksekgonul2024textgrad,yang2024large}, or embedding-level optimization directions~\cite{lester2021power,li2021prefix}, but provide limited insight into the edit-level behaviors underlying optimizer failures. As a result, current evaluations often reveal \emph{whether} optimization succeeds, but not \emph{which prompt modifications} systematically contribute to improvement or degradation across different task settings. Yet understanding these edit-level behaviors is important for both diagnosis and optimizer design: two optimizers may achieve similar average gains while relying on fundamentally different editing strategies, and seemingly beneficial prompt modifications may interact differently with different reasoning tasks.



\begin{figure}[t]
  \centering
  \includegraphics[width=\columnwidth]{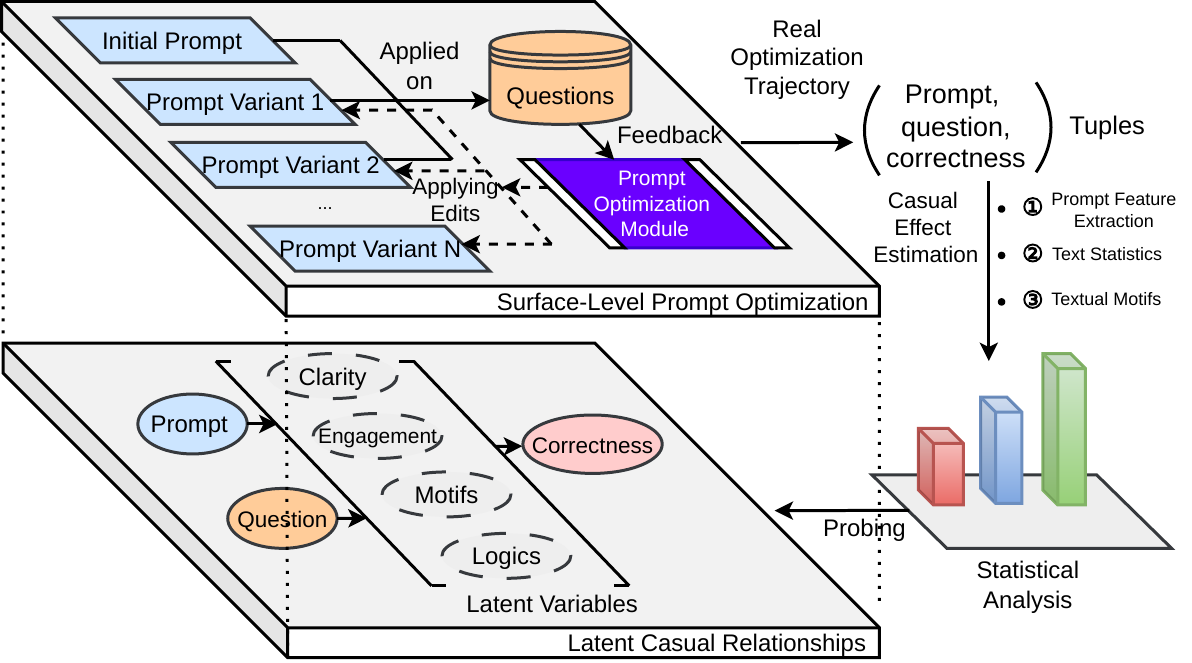}
    \caption{
    Overview of our multi-view probing framework for prompt optimizer behavior. 
    }
  \label{fig:figure}
\end{figure}

In this work, we investigate prompt optimizer behavior through an observational multi-view analysis of optimizer-induced prompt edits, as illustrated in Figure~\ref{fig:figure}. Rather than treating optimized prompts as indivisible artifacts, we analyze consecutive prompt revisions within real optimization trajectories and examine how different edit families are associated with downstream performance changes across task groups. 

To reduce dependence on any single representation of prompt edits, we probe optimizer behavior through three complementary views: (1) GPT-4o-annotated cognitive and instructional features, (2) deterministic surface-level text statistics, and (3) literal text-diff motifs extracted from consecutive prompt revisions. These complementary representations are subsequently integrated into a propensity-adjusted associational analysis framework to characterize heterogeneous optimizer behaviors across reasoning tasks.

Methodologically, our analysis adopts a causal inference-inspired observational framework. We use propensity-adjusted associational estimation~\cite{rosenbaum1983central} to reduce measured selection bias arising from differences in prior prompt states, while explicitly avoiding strong causal claims. Because the analysis involves many simultaneous feature-task comparisons, we distinguish between statistically robust findings that survive false-discovery-rate (FDR) correction and exploratory directional patterns that provide corroborative but non-confirmatory evidence. Throughout the paper, we therefore organize results using a unified two-tier evidence hierarchy. \textit{Tier 1} refers to associations surviving Benjamini--Hochberg false discovery rate correction. \textit{Tier 2} refers to directionally consistent corroborative patterns reproduced across multiple representations or frameworks but not surviving full multiple-testing correction. 


Across around 20 thousand real world prompt optimization's (\textit{prompt}, \textit{question}, \textit{correctness}) tuples, we observe consistent edit-level heterogeneity across task groups. In particular, complexity-increasing and meta-instructional edits tend to be negatively associated with mathematical and multi-hop reasoning performance, whereas metacognitive and step-structured edits are positively associated with logical and sequential reasoning tasks. Several of these associations remain significant after false-discovery-rate correction, while others appear consistently across multiple independent prompt representations.

Our main contributions are summarized as follows:
\begin{itemize}[leftmargin=*]

\item We present an observational edit-level analysis of prompt optimizer behavior across multiple optimization frameworks, LLM backbones, and reasoning task groups.

\item We identify statistically robust heterogeneous associations between prompt optimizer-induced edits and downstream task performance. 


\item We provide evidence that prompt optimization failures are not purely random, but are systematically associated with interactions between edits and benchmark-specific characteristics, motivating future task-conditioned optimizer design.

\end{itemize}

\section{Related Work}
\label{sec:related}

\paragraph{Automated prompt optimization.}
Automated prompt optimization treats the prompt as a learnable variable to maximize task performance.
Early work learns soft prompts or searches discrete prompt tokens/templates for few-shot adaptation and probing~\citep{li2021prefix,lester2021power,shin2020autoprompt,gao2021making}.
Later black-box methods optimize natural-language instructions directly, including instruction generation and ranking~\citep{zhou2022large}, gradient-free edit search~\citep{prasad2023grips}, reinforcement-learning-based prompt editing~\citep{deng2022rlprompt,zhang2022tempera}, and textual-gradient revision~\citep{pryzant2023automatic,yuksekgonul2024textgrad}.
More recent systems use LLMs as optimizers or search operators, through natural-language proposal selection~\citep{yang2024large}, evolutionary/self-referential mutation~\citep{guo2024connecting,fernando2023promptbreeder,gepa}, and program-level prompt compilation~\citep{khattab2024dspy,opsahlong2024miprov2}.
Despite their different optimization mechanisms, these methods converge to a common empirical failure mode in our setting: consistent performance degradation on math and multi-hop tasks.
This convergence motivates a feature-level diagnosis of prompt changes rather than a method-specific analysis.

\paragraph{Prompt sensitivity and format effects.}
A complementary line of work studies how non-semantic prompt properties affect LLM performance.
\citet{zhao2021calibrate} shows that few-shot calibration can substantially reduce format-induced bias, while \citet{lu2022ordering} finds that demonstration order can change accuracy by up to 30 points.
Related studies further show that in-context learning depends on label space, input distribution, and sequence format more than exact input-label mappings~\citep{min2022rethinking}, that models can be insensitive to instruction semantics~\citep{webson2022prompt}, that answer surface forms distort likelihood-based scoring~\citep{holtzman2021surface}, and that minor formatting choices can induce large performance swings~\citep{sclar2024quantifying}.
Together, these findings demonstrate that prompt surface properties matter independently of semantic content.
We extend this perspective from isolated format variation to optimizer trajectories, examining consecutive optimizer-induced edits and their task-type-conditioned performance effects.
The Coin Flip paper~\citep{zhang2026coinflip} uses ANOVA-based analysis to identify run-level conditions under which optimization succeeds; we complement this with feature-level and edit-level causal analysis across task types.

\paragraph{Causal inference for NLP.}
Causal inference has been increasingly used in NLP for debiasing, model behavior analysis, and text-based causal estimation~\citep{feder2022causal}.
Recent LLM-era work further studies LLMs both as objects of causal analysis and as tools for causal discovery or effect estimation~\citep{ma2025causal,liu2025large}.
For example, recent studies use LLMs to estimate causal effects from unstructured text~\citep{dhawan2024end}, build causal graphs and perform counterfactual inference from natural language~\citep{gendron2024counterfactual}, generate or match counterfactual texts for faithful model explanation~\citep{gat2023faithful,wang2024beyond}, and evaluate LLMs' formal causal and counterfactual reasoning abilities~\citep{jin2024can,maasch2025compositional,chen2026counterbench}.
The heterogeneous treatment effect literature~\citep{athey2018wager} provides the statistical framework for conditional causal effects across subpopulations, while the double machine learning framework~\citep{chernozhukov2018double} enables nuisance-adjusted effect estimation with flexible machine learning models.
We adapt inverse probability of treatment weighting (IPTW)~\citep{robins2000marginal} to prompt optimization traces, where the treatment is an optimizer-induced change in prompt features and the confounder is the prior prompt state.
This framing extends causal-effect estimation tools to optimizer behavior analysis, while our evidence-tier design distinguishes FDR-controlled associations from exploratory corroboration.

\paragraph{Closest prior work.}
CPO~\citep{chen2026cpo} applies double machine learning to whole-prompt embeddings to estimate optimization effects, whereas our focus is diagnostic and feature-level: we analyze how interpretable prompt edits are associated with gains or losses across task groups. While the CausalNLP survey~\citep{feder2022causal} reviews text-as-treatment methods, we instantiate this perspective at the optimizer edit-motif level. Recent work on embedding-based causal inference~\citep{dawoud2026reading} shows dense representations can reduce selection bias; in contrast, we emphasize interpretable surface features that support direct comparison across annotation-based and annotation-free views. The Coin Flip paper~\citep{zhang2026coinflip} studies optimizer success at the run level via ANOVA, whereas we perform edit-level IPTW-adjusted analysis to identify which specific edit types are associated with optimizer success or failure.

\paragraph{Cognitive load in instructional design.}
Our annotation scheme draws on cognitive load theory, which distinguishes intrinsic load from task complexity, extraneous load from irrelevant or poorly structured information, and germane load from task-relevant cognitive processing. We use GPT-4o to annotate these constructs at the prompt level and validate them against deterministic text proxies (\S\ref{sec:annotation}), following partial-validity assessment practices in computational social science.
\section{Data and Annotation}
\label{sec:data}

\subsection{Pairwise Prompt Comparison Dataset}

We collected optimization logs from 3 frameworks, i.e., DSPy~\cite{khattab2024dspy} (with MiPROv2 optimizer~\cite{opsahlong2024miprov2}), TextGrad~\cite{yuksekgonul2024textgrad}, and GEPA~\cite{gepa}; 5 LLM backbones, i.e., GPT-5.2, GPT-4o, Qwen3-32B~\cite{yang2025qwen3}, Deepseek-v3, and Deepseek-R1~\cite{liu2024deepseek}; and 11 NLP benchmarks spanning 5 reasoning task categories:

\noindent\circleone{1} Commonsense reasoning (including Commonsense QA~\cite{talmor2019commonsenseqa}, causal judgment, disambiguation QA datasets~\cite{suzgun2023bbh}, with a mean base accuracy 0.70), \\
\noindent\circletwo{2} Mathematical reasoning (including GSM8K~\cite{cobbe2021gsm8k}, MultiArith datasets~\cite{roy2015solving}, with a mean base accuracy 0.97), \\
\noindent\circlethree{3} Logical reasoning (including boolean expressions, coinflip~\cite{suzgun2023bbh} datasets, with a mean base accuracy 0.81), \\
\noindent\circlefour{4} Sequential reasoning (including last letters~\cite{suzgun2023bbh}, ListOps~\cite{keysers2019measuring} datasets, with a mean base accuracy 0.75), \\
\noindent\circlefive{5} Multi-hop reasoning (including Strategy QA~\cite{geva2021did}, date understanding~\cite{suzgun2023bbh} datasets, with a mean base accuracy 0.75). 

Especially, Math benchmarks exhibit particularly high initial performance,raising the possibility of a ceiling effect that we further examine in \S\ref{sec:cate}.

For each consecutive optimization step, we record the performance gain, defined as $\gain = \mathrm{acc}(p_2) - \mathrm{acc}(p_1)$,  on a fixed evaluation set. $p_1$ and $p_2$ are the former and updated prompts, and the prompt optimization applies the edit on $p_1$ to produce $p_2$.
This process yields 2,095 pairwise comparisons from DSPy for the main analysis, and an additional 17,708 comparisons from TextGrad and GEPA for cross-framework replication.

\subsection{Prompt Feature Annotation}
\label{sec:annotation}

Each prompt is annotated with 12 features derived from cognitive load theory and instructional design, each scored on a  1--10 scale by GPT-4o. 

\begin{table}[h]
\centering
\small
\resizebox{\columnwidth}{!}{%
\begin{tabular}{ll}
\toprule
\textbf{Feature} & \textbf{Description} \\
\midrule
Clarity & Precision/unambiguity of task specification \\
Engagement & Motivational framing \\
Politeness & Tone politeness level \\
Intrinsic\_load & Core task complexity \\
Extraneous\_load & Redundant/irrelevant information \\
Enc.\_germane\_load & Prompts for effortful processing \\
Objectives & Explicit goal specification \\
Metacognition & Self-monitoring or reasoning prompts \\
Demos & Few-shot demonstration count/quality \\
Structural\_logic & Logical organization and sequencing \\
Contextual\_logic & Context coherence \\
Hallucination\_aw. & Uncertainty acknowledgment \\
\bottomrule
\end{tabular}}
\caption{The 12 GPT-4o-annotated cognitive-load features.}
\label{tab:features}
\end{table}

\paragraph{Construct validity.}
We correlate these annotations against 13 text-derived features (\S\ref{sec:surface}) using Spearman rank correlation.
7 of 19 expected correlations are validated ($|\rho| > 0.10$, $p < 0.10$): Metacognition $\leftrightarrow$ meta-cognitive word density ($\rho = +0.691$, $p < 0.001$); Intrinsic\_load $\leftrightarrow$ word count ($\rho = +0.566$, $p < 0.001$); Structural\_logic $\leftrightarrow$ numbered list presence ($\rho = +0.457$, $p < 0.001$).
Annotation validity is partial but sufficient for coarse-grained analysis, and we address the remaining validity concern through annotation-free replication in \S\ref{sec:textnative}.

\section{Confounding-Adjusted Associational Analysis}
\label{sec:cate}

\subsection{Estimand and Assumptions}
\label{sec:estimand}

We estimate Inverse Probability of Treatment Weighting~\cite{rosenbaum1983central} (\emph{IPTW)-adjusted conditional mean gain differences} (ACMGD) for optimizer-selected edit regimes. ACMGD is therefore interpreted as an IPTW-adjusted observational association rather than as an interventional causal estimand. For a given feature indicator $T$, we set treatment $= 1$ when the revision from $p_1$ to $p_2$ increases that feature, and $0$ otherwise. Within each dataset/task group, the estimand compares expected performance gain between edit regimes with and without $T$ , under three standard assumptions:

\begin{itemize}[leftmargin=*,topsep=2pt,itemsep=1pt]
\item \textbf{Consistency}: The observed outcome equals the potential outcome under the observed treatment.
\item \textbf{Positivity}: Each unit has positive probability of both treatment values given covariates.
\item \textbf{Conditional exchangeability}: Given \emph{pre-edit covariates} (prompt~$p_1$ state: length, demo count, headroom, framework, backbone, dataset, optimization step), treatment assignment is conditionally independent of untreated potential outcomes.
\end{itemize}

\paragraph{Plausible unmeasured confounders.}
Conditional exchangeability is unlikely to hold exactly in this setting.
Several optimizer-internal variables remain unmeasured: (1)~\emph{search trajectory history} may confound the estimates, because optimizers tend to insert meta-instructions after observing prior failures, leaving ``meta-instruction inserted'' correlated with ``prior prompt was struggling''; (2)~\emph{selection pressure} may arise because high-performing prompts at later optimization steps are more likely to receive additional complex features; (3)~\emph{LLM-specific instruction following} may matter because backbones vary in how they process instruction-level directives, and backbone is partially conditioned on rather than fully controlled.
IPTW adjusts for measured pre-edit state but cannot eliminate these residual dependencies.
Accordingly, the results should be interpreted as \emph{adjusted associational contrasts} rather than causal effects.

\paragraph{Treatment bundling.}
Prompt optimizers often modify multiple features simultaneously, so treatments are \emph{edit bundles} rather than isolated feature changes. Although the propensity model conditions only on pre-edit covariates, residual bundle heterogeneity remains a limitation.
Thus, the analysis provides \emph{observational associational evidence rather than causal identification}.

\subsection{Method: IPTW within Task Types}
\label{sec:method}



For each feature $\feat$ and task group $\tasktype$, we proceed as follows.
First, we define treatment indicator as
$T = \mathbf{1}[\feat\text{-change} > 0]$.
Second, we fit a logistic propensity model
$P(T{=}1 \mid X)$ using only pre-edit covariates.
Third, we compute stabilized IPTW weights,
\[
w_i =
\frac{T_i \,\bar{P}(T{=}1)}{e(X_i)}
+
\frac{(1-T_i)\,\bar{P}(T{=}0)}{1-e(X_i)},
\]
with weights capped at 10 to limit the influence of extreme propensity scores.

Fourth, we estimate the stabilized-IPTW weighted
average conditional mean gain difference (ACMGD),
\[
\widehat{\mathrm{ACMGD}}_{\feat,\tasktype}^{\mathrm{SIPW}}
=
\frac{\sum_i w_i T_i Y_i}{\sum_i w_i T_i}
-
\frac{\sum_i w_i (1-T_i) Y_i}{\sum_i w_i (1-T_i)},
\]
where $Y_i$ denotes the observed performance gain for revision pair $i$.

Fifth, we calculate block-bootstrap standard errors  using 500 resamples, with blocks defined by dataset $\times$ backbone stratum to account for within-run and within-dataset correlation. All reported p-values and BH-corrected significance stars are based on these block-bootstrap uncertainties. BH-FDR correction is applied simultaneously across all 60 (feature $\otimes$ task-group) tests.

The average confounding magnitude across all combinations is $|\sipw - \text{naive}| = 0.018$, confirming that selection bias is present in naive estimates.

\subsection{Results}
\label{sec:cate_results}

We organize results into two tiers: \textbf{Tier 1 (Confirmatory)} contains BH-FDR corrected findings ($q < 0.05$), whereas \textbf{Tier 2 (Exploratory)} contains uncorrected directional patterns that corroborate Tier 1 but do not independently support inference.

\paragraph{Tier 1: Confirmatory findings (BH-FDR corrected).}
Table~\ref{tab:cate_main} reports IPTW-adjusted ACMGD estimates by feature and dataset/task group.
Across 60 simultaneous tests, 11 reach uncorrected significance ($p < 0.05$) and \textbf{2 survive Benjamini--Hochberg FDR correction} ($q < 0.05$):

\begin{itemize}[leftmargin=*,topsep=2pt,itemsep=2pt]
\item \textbf{Extraneous\_load $\otimes$ sequential reasoning}: ACMGD $= -0.060^{\bigstar}$\footnote{$\bigstar$~=~BH-FDR $q{<}0.05$; $^*$~=~uncorrected $p{<}0.05$. These significance markers are used consistently throughout the paper. More details are in Appendix~\ref{app:fdr}.}.
Increases in extraneous load are significantly associated with performance degradation in the sequential task group (LOO-stable across 5/5 splits). Here, leave-one-out (LOO) stability refers to re-estimating the sign and direction of an association after iteratively excluding one dataset at a time, providing a robustness check against single-dataset-driven effects.
\item \textbf{Metacognition $\otimes$ sequential reasoning}: ACMGD $= +0.062^{\bigstar}$.
Metacognitive prompting is significantly associated with performance gains in the sequential task group (LOO-stable across 2/3 splits; one flip in disambiguation\_qa).
\end{itemize}

Per-dataset robustness details for the Tier 1 findings are provided in Appendix~\ref{app:tier1_robustness}.

\begin{figure}[t]
  \centering
  \includegraphics[width=\columnwidth]{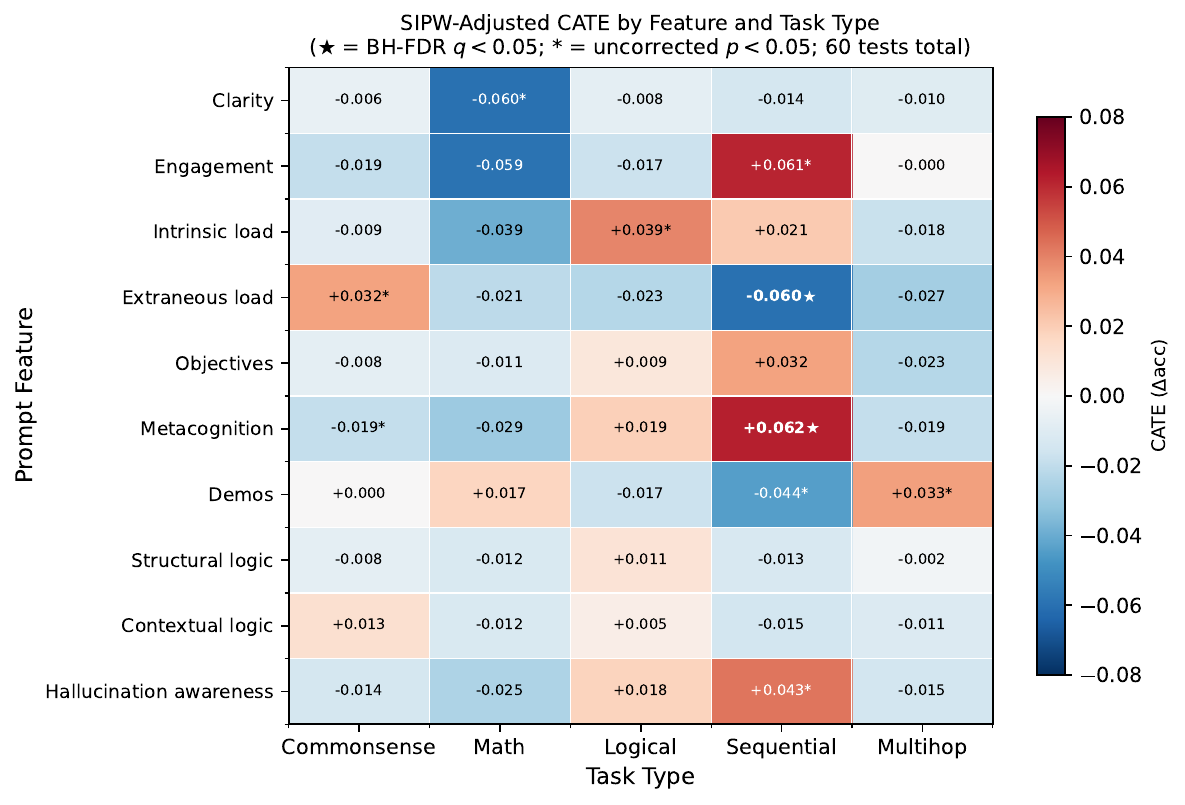}
  \caption{IPTW-adjusted ACMGD heatmap (10 features $\times$ 5 dataset/task groups, 60 tests).
  Blue = positive association; red = negative. 
  Math and multihop task groups show predominantly negative associations with complexity-increasing features; this directional pattern is exploratory (uncorrected).}
  \label{fig:cate_heatmap}
\end{figure}

\begin{table}[t]
\centering
\footnotesize
\setlength{\tabcolsep}{3pt}
\resizebox{\columnwidth}{!}{%
\begin{tabular}{lrrrrrrr}
\toprule
\textbf{Feature} & \textbf{CS} & \textbf{Math} & \textbf{Logic} & \textbf{MH} & \textbf{Seq} & \textbf{Spread} \\
\midrule
Clarity             & $-0.006$ & $-0.060^*$ & $-0.008$ & $-0.010$ & $-0.014$ & 0.053 \\
Engagement          & $-0.019$ & $-0.059$   & $-0.017$ & $-0.000$ & $+0.061^*$ & 0.120 \\
Intrinsic\_load     & $-0.009$ & $-0.039$   & $+0.039^*$ & $-0.018$ & $+0.021$ & 0.078 \\
\textbf{Extraneous\_load} & $+0.032^*$ & $-0.021$ & $-0.023$ & $-0.027$ & $\mathbf{-0.060}^{\bigstar}$& \textbf{0.092} \\
Objectives          & $-0.008$ & $-0.011$ & $+0.009$ & $-0.023$ & $+0.032$ & 0.055 \\
\textbf{Metacognition} & $-0.019^*$ & $-0.029$ & $+0.019$ & $-0.019$ & $\mathbf{+0.062}^{\bigstar}$ & \textbf{0.091} \\
Demos               & $+0.000$ & $+0.017$ & $-0.017$ & $+0.033^*$ & $-0.044^*$ & 0.077 \\
Structural\_logic   & $-0.008$ & $-0.012$ & $+0.011$ & $-0.002$ & $-0.013$ & 0.024 \\
Contextual\_logic   & $+0.013$ & $-0.012$ & $+0.005$ & $-0.011$ & $-0.015$ & 0.027 \\
Hallucination\_aw.  & $-0.014$ & $-0.025$ & $+0.018$ & $-0.015$ & $+0.043^*$ & 0.067 \\
\bottomrule
\end{tabular}}
\caption{IPTW-adjusted ACMGD by feature and task group. CS=commonsense, Math=Mathematical, Logic=logical, Seq=sequential, MH=multihop, Spread = max$-$min.}
\label{tab:cate_main}
\end{table}

\paragraph{Tier 2: Exploratory directional patterns.}
Beyond the two BH-corrected effects, several uncorrected patterns appear directionally consistent and corroborate Tier 1:
the Extraneous\_load sign reversal (commonsense $+0.032^*$ vs.\ sequential $-0.060^{\bigstar}$) is stable in 5/5 LOO splits.
Demos (math $+0.017$ vs.\ sequential $-0.044^*$) is directionally consistent with the surface feature analysis in \S\ref{sec:textnative}. In addition, math task groups show predominantly negative point estimates for complexity-increasing features across 9/12 features, a pattern consistent with the motif analysis but uncorrected. These results are reported as \emph{exploratory hypothesis-generating observations only} and do not independently support inferential claims.

\subsection{Ceiling Effects}
\label{sec:ceiling}

We examine whether dataset/task-group heterogeneity can be explained by ceiling effects alone.
Headroom. defined as $1 - \text{base\_accuracy}$, captures the remaining potential for improvement.
The association between headroom and performance gain is weak and not statistically significant, with Spearman $\rho(\text{headroom}, \gain) = 0.047$ ($p = 0.053$). In addition, the partial $R^2$ of task group conditional on headroom is small but nonzero, $R^2(\text{task group} \mid \text{headroom}) = 0.003$, indicating that task-group membership explains residual variation beyond headroom.
These results suggest that ceiling effects are only a minor contributor and that dataset/task-group moderation is not reducible to baseline performance differences alone.

\section{Annotation-Free Corroboration}
\label{sec:textnative}

To assess whether the BH-corrected findings in \S\ref{sec:cate} depend on GPT-4o annotation choices, we re-examine the same prompt pairs using annotation-free text representations.
These analyses produce Tier 2 corroborating: they align with Tier 1 findings but are not independently confirmatory.

\subsection{Surface Complexity Features}
\label{sec:surface}

We extract 14 deterministic text features from raw instruction text, including word count, demonstration count extracted from the JSON structure, step-word density based on terms such as “step,” “first,” and “then,” type-token ratio, compression ratio, sentence count, and related surface statistics.
We then apply the same IPTW-adjusted associational analysis on DSPy pairs ($N = 2{,}095$).

\begin{table}[t]
\centering
\footnotesize
\setlength{\tabcolsep}{3pt}
\resizebox{\columnwidth}{!}{%
\begin{tabular}{llrrrrrrr}
\toprule
\textbf{Feature} & \textbf{LLM proxy} & \textbf{CS} & \textbf{Math} & \textbf{Log} & \textbf{Seq} & \textbf{MH} & \textbf{Spread} \\
\midrule
\textbf{n\_demos} & Demos & $+0.011$ & $\mathbf{+0.320^*}$ & $+0.081^*$ & $\mathbf{-0.043^*}$ & $\mathbf{+0.238^*}$ & \textbf{0.362} \\
word\_count & Extr.\ load & $-0.009$ & $-0.009$ & $+0.051^*$ & $\mathbf{+0.097^*}$ & $\mathbf{-0.060^*}$ & 0.157 \\
type\_tok.\ ratio & Clarity & $-0.016$ & $\mathbf{-0.091^*}$ & $+0.024$ & $+0.059^*$ & $+0.001$ & 0.151 \\
\textbf{step\_words} & Struct.\ logic & $-0.006$ & $-0.044$ & $\mathbf{+0.045^*}$ & $\mathbf{+0.075^*}$ & $\mathbf{-0.051^*}$ & \textbf{0.125} \\
reasoning\_words & Metacognition & $-0.023$ & $-0.065$ & $\mathbf{+0.070^*}$ & $\mathbf{+0.064^*}$ & $-0.013$ & 0.135 \\
compr.\ ratio & --- & $-0.000$ & $-0.058^*$ & $-0.053^*$ & $-0.043^*$ & $+0.010$ & 0.068 \\
\bottomrule
\end{tabular}}
\caption{Surface feature IPTW-adjusted ACMGD. CS=commonsense, Math=Mathematical, Logic=logical, Seq=sequential, MH=multihop, Spread = max$-$min.}
\label{tab:surface_cate}
\end{table}

All four major LLM-annotated directional patterns are reproduced by text-native features (Table~\ref{tab:surface_cate}). Specifically, n\_demos algins with Demos (math positive, sequential negative);
type\_token\_ratio algins with Clarity (math negative);
step\_words algins with Structural\_logic (logical/sequential positive, multihop negative);
word\_count algins with Extraneous\_load (sequential positive, multihop negative).

\begin{figure*}[tbh]
  \centering
  \includegraphics[width=\textwidth]{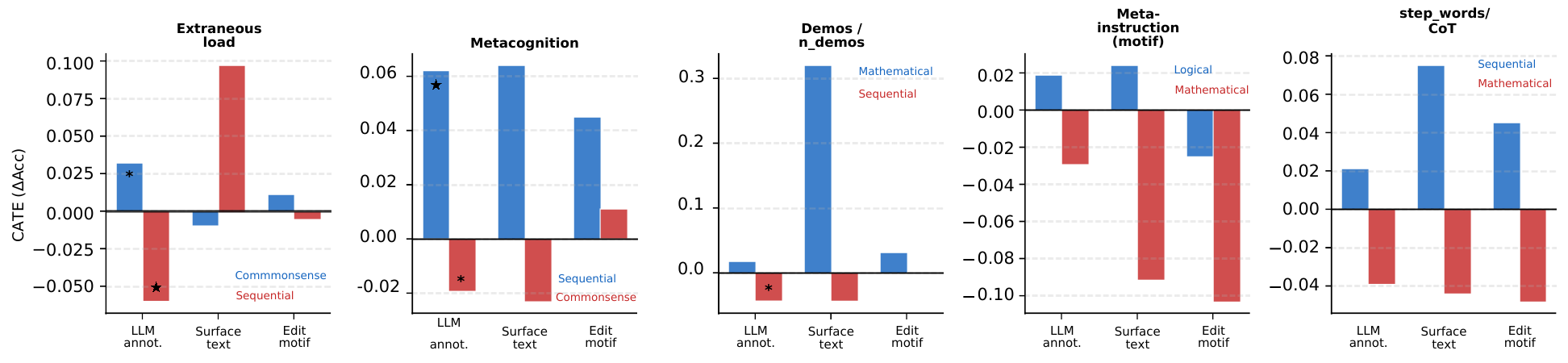}
  \caption{Multi-view convergence of sign reversals across three representations (LLM annotation, surface text features, edit motifs) for five key feature comparisons.
  Blue dots = higher-benefit task type; red dots = lower-benefit task type.
  The consistent sign pattern across representations reduces concern that results are annotation artifacts.}
  \label{fig:convergence}
\end{figure*}

This convergence reduces the concern that the observed patterns are artifacts of GPT-4o annotation choices.
If the sign reversals were driven solely by GPT-4o annotation choices, deterministic surface proxies would be unlikely to reproduce the same directional patterns.
Although this comparison does not fully rule out annotation bias, becauseboth representations are derived from the same underlying prompts,it provides annotation-independent corroboration.
Figure~\ref{fig:convergence} visualizes this multi-view agreement for five key feature comparisons.

\paragraph{Cross-framework directional corroboration (TextGrad, GEPA, $N = 17{,}708$).}
We apply the same surface-feature extraction to TextGrad and GEPA pairs as a directional corroboration check.
Because TextGrad and GEPA uses a different optimization framework and prompt structure, this analysis is a \emph{partial replication of directional patterns only}, rather than a full replication of the IPTW adjustment protocol.
The results show consistent directional pattern:
word\_count (sequential $+0.017^*$, multihop $-0.014^*$);
sentence\_count (commonsense $+0.027^*$ vs.\ sequential $-0.021^*$ vs.\ multihop $-0.018^*$);
reasoning\_words (sequential $+0.039^*$ vs.\ commonsense $-0.014^*$).
The n\_demos feature is less interpretable for TextGrad (which does not use structured few-shot demos in the same format).
We therefore interpret these results as cross-framework directional corroboration, instead of independent replication.

\subsection{Edit Motif Effects}
\label{sec:motifs}

We compute text-diff for each consecutive prompt pair, extract inserted word spans of at least $\geq 4$ tokens, and classify them into four pre-specified motif categories using regex patterns defined prior to analysis:
\textit{chain\_of\_thought} (``step by step'', ``think through'', ``reasoning'', etc.),
\textit{meta\_instruction} (``make sure to'', ``do not'', ``ensure'', ``remember to'', etc.),
\textit{step\_by\_step} (``step 1'', ``first...then'', numbered instruction lists),
and \textit{clarity\_constraint} (``concisely'', ``briefly'', ``simple'', ``avoid'', etc.).

We then estimate IPTW-adjusted ACMGD for each motif $\times$ dataset/task-group combination using pre-treatment covariates, and apply BH-FDR correction across all 15 motif $\times$ group tests simultaneously.

\begin{table*}[t]
\centering
\small
\setlength{\tabcolsep}{4pt}
\begin{tabular}{lrrrrrr}
\toprule
\textbf{Motif} & \textbf{CS} & \textbf{Math} & \textbf{Logic} & \textbf{Seq} & \textbf{MH} & \textbf{spread} \\
\midrule
chain\_of\_thought       & $+0.011$ & $-0.048$ & $+0.026$ & $+0.045^*$ & $+0.031$ & 0.093 \\
\textbf{meta\_instruction} & $+0.011$ & $\mathbf{-0.103}^{\bigstar}$ & $-0.025$ & $-0.005$ & $-0.044^*$ & \textbf{0.114} \\
\textbf{clarity\_constraint} & $-0.011$ & $+0.000$ & $\mathbf{-0.083}^{\bigstar}$ & $+0.011$ & $+0.012$ & \textbf{0.094} \\
step\_by\_step           & $+0.000$ & $+0.031$ & $+0.000$ & $+0.000$ & $+0.000$ & 0.031 \\
\bottomrule
\end{tabular}
\caption{Edit motif insertion ACMGD by dataset/task group. 
CS=commonsense, Math=Mathematical, Logic=logical, Seq=sequential, MH=multihop, Spread = max$-$min.Motif co-occurrence: chain\_of\_thought \& meta\_instruction in 24.0\% of pairs. Results organized by Tier 1 (starred) vs.\ Tier 2 (unstated).}
\label{tab:motif_cate}
\end{table*}

\begin{figure}[tbh]
  \centering
  \includegraphics[width=\columnwidth]{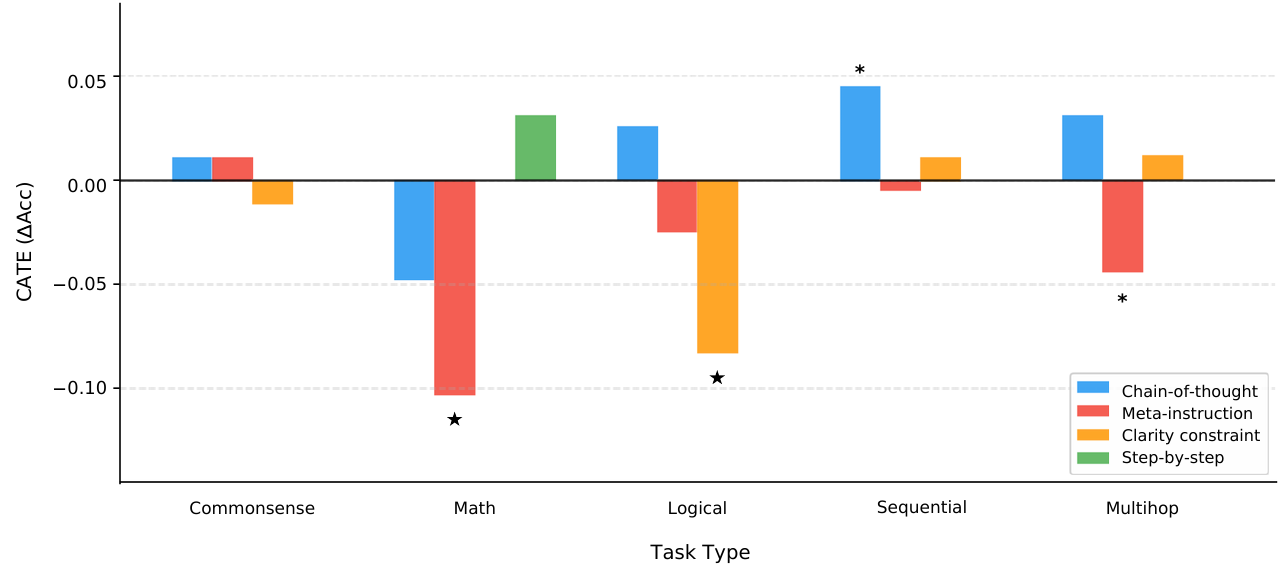}
  \caption{Edit motif insertion ACMGD by dataset/task group (4 motifs $\times$ 5 groups, 15 tests). Meta-instruction is significantly associated with worse math performance; clarity-constraint is significantly associated with worse logical performance.}
  \label{fig:motif_cate}
\end{figure}

\paragraph{Tier 1: BH-corrected motif findings.}

\textit{Meta-instruction} insertion is significantly associated with lower performance in the math dataset group (ACMGD~$= -0.103^{\bigstar}$, BH-corrected).
\emph{Per-dataset caveat}: This effect is partially concentrated in MultiArith (naive CATE~$= -0.277$, $n = 64$) vs.\ GSM8K ($-0.019$, $n = 14$); the aggregated math estimate should be interpreted with this per-dataset heterogeneity in mind (see Appendix~\ref{app:motif_dataset}).
Meta-instruction is also directionally negative in multihop ($-0.044^*$, uncorrected, Tier 2).

\textit{Clarity-constraint} insertion is significantly associated with lower performance in the logical dataset group (ACMGD~$= -0.083^{\bigstar}$, BH-corrected).
Instructions to ``be concise/brief/simple'' may suppress step-by-step reasoning on boolean and coin-flip tasks.

\paragraph{Tier 2: Exploratory directional pattern.}

\textit{Chain-of-thought} shows a preserved directional pattern across representations: negative in math groups (motif: $-0.048$; surface step\_words: $-0.044$; LLM Structural\_logic negative), positive in sequential/last\_letters (motif: $+0.045^*$; surface step\_words: $+0.075^*$; LLM Structural\_logic positive).
This pattern is consistent but uncorrected across all representations and should be treated as Tier 2 (exploratory corroboration).
\emph{Dataset caveat}: ``sequential'' is a single-dataset group (last\_letters); this is dataset-level evidence, not multi-dataset task-type evidence.

\paragraph{Motif validity.}
We manually audited 50 classified-positive spans per key motif to assess precision.
For \textit{meta\_instruction}, representative inserted spans include:
\begin{itemize}[leftmargin=*,topsep=1pt,itemsep=1pt]
  \item ``\textit{Ensure that both outputs clearly explain how the final solution was reached based on the arithmetic operations involved.}''
  \item ``\textit{Make sure to articulate the reasoning process clearly, even if it doesn't require detailed breakdowns.}''
\end{itemize}
These patterns are consistently genuine meta-instructional directives added by the optimizer.
For \textit{chain\_of\_thought}, the most common false positive is the JSON key \texttt{"reasoning": "..."} appearing in structured few-shot demo fields, i.e., the regex matches the word ``reasoning'' but the pattern appears in a non-instruction field not visible to the model as a directive.
This false positive inflates the treatment indicator for chain\_of\_thought without representing an actual user-visible instruction change.
We estimate meta\_instruction precision at ${\gtrsim}90\%$ based on this audit (50 random positive spans, stratified by dataset; labeled independently by one author; ambiguous cases counted as false positives).
Chain\_of\_thought precision is lower and the pattern should be treated as noisier.
Motif precision was not formally estimated at scale and the audit involved one labeler; results remain preliminary.

\section{Conclusion}
\label{sec:conclusion}

We present a multi-view observational analysis of edit-level heterogeneity in prompt optimizer behavior.
Using three complementary representations of 2,095 pairwise prompt comparisons, i.e., GPT-4o-annotated cognitive-load features, deterministic surface text statistics, and literal text-diff motifs, we find four BH-FDR-corrected associations: Extraneous\_load $\otimes$ sequential ($-0.060^{\bigstar}$), Metacognition $\otimes$ sequential ($+0.062^{\bigstar}$), meta-instruction $\otimes$ math ($-0.103^{\bigstar}$), and clarity-constraint $\otimes$ logical ($-0.083^{\bigstar}$).
All four involve at least two benchmark datasets per task group. Beyond these confirmatory findings, directionally consistent corroborating evidence appears in surface text features and in 17,708 TextGrad/GEPA pairs, and 83\% of headline sign patterns survive LOO exclusion.

These patterns suggest that edit-family-level analysis, especially when conditioned on benchmark-specific task demands, offers a more informative diagnosis than aggregate success or failure metrics alone.
The pre-optimization question ``does my target benchmark resemble groups where meta-instruction and complexity-increasing edits are negatively associated with gains?'' has practical diagnostic value even under the observational limitations of this study, and points toward edit-conditioned optimizer design as a fruitful direction for interventional follow-up.

\paragraph{Implications. }
For practitioners, the ACMGD tables in this paper serve as a preliminary diagnostic: before running an expensive optimizer, consider whether the target task/dataset resembles groups (math-like, multihop-like) where meta-instruction and complexity-increasing edits show negative associations with gains.
This does not guarantee failure, but the BH-corrected associations are consistent across frameworks, representations, and LOO folds. The consistent negative association between meta-instruction insertion and math-type benchmarks suggests that future prompt optimizers may benefit from task-conditioned edit control, particularly by discouraging unnecessary complexity-increasing or meta-instructional edits during optimization.


\paragraph{Reproducibility.}
All analysis code and pairwise comparison dataset will be released upon publication.
The TextGrad processing pipeline is compatible with TextGrad public official repository.

\section*{Limitations}
\label{sec:limitations}

Our analysis is observational rather than interventional. Although IPTW reduces measured selection bias using pre-edit prompt states, residual confounding may remain due to optimizer trajectory history, backbone-specific instruction following, and bundled prompt edits that modify multiple features simultaneously. Accordingly, the reported ACMGD estimates should be interpreted as adjusted associational contrasts rather than causal effects.

In addition, several findings remain benchmark-sensitive. Some task-group effects are partially concentrated in specific datasets (e.g., MultiArith within math), and cross-framework corroboration is directional rather than a fully controlled replication because different optimization frameworks use different prompt structures. Finally, motif extraction relies on pre-specified regex patterns and therefore introduces unavoidable labeling noise, particularly for chain-of-thought-related edits.

\bibliography{reference}

\appendix
\section{Statistical Significance (Uncorrected \textit{p}-values, BH-FDR Correction)}
\label{app:fdr}

Our analysis involves multiple simultaneous hypothesis tests across prompt features, edit motifs, and dataset/task groups. In such settings, interpreting raw statistical significance without correction can substantially inflate the probability of false discoveries. This appendix clarifies the distinction between uncorrected \textit{p}-values and Benjamini--Hochberg false discovery rate (BH-FDR) corrected results used throughout the paper.

\subsection{Uncorrected \textit{p}-values}

For each feature--task-group combination, we estimate an IPTW-adjusted associational effect (ACMGD) and compute a corresponding hypothesis test:
\begin{equation}
H_0 : \mathrm{ACMGD} = 0,
\end{equation}
where the null hypothesis assumes no adjusted association between the optimizer-induced edit regime and performance gain.

The reported uncorrected \textit{p}-value measures the probability of observing an effect at least as extreme as the estimated one under the null hypothesis:
\begin{equation}
p = \Pr\left(|T| \ge |T_{\mathrm{obs}}| \mid H_0 \right),
\end{equation}
where $T$ denotes the test statistic computed using block-bootstrap uncertainty estimates.

Throughout the paper, an uncorrected significance marker
\[
{}^{*} : p < 0.05
\]
indicates that the corresponding association is statistically significant when considered as an isolated test.

However, because our analysis evaluates many hypotheses simultaneously (60 feature $\times$ task-group tests in Table~2 and 15 motif $\times$ group tests in Table~4), interpreting uncorrected \textit{p}-values alone would inflate the probability of false positives due to the multiple comparisons problem.

\subsection{Multiple Comparisons Problem}

Suppose $m$ independent null hypotheses are tested at significance level $\alpha = 0.05$. Even if all null hypotheses are true, the expected number of false positives is:
\begin{equation}
m \times \alpha.
\end{equation}

For example, with $m=60$ tests, approximately
\[
60 \times 0.05 = 3
\]
false discoveries are expected purely by chance under naive thresholding.

Consequently, some apparently significant associations may arise from stochastic variation rather than systematic optimizer behavior.

\subsection{Benjamini--Hochberg False Discovery Rate Correction}

To mitigate inflated false positives, we apply the Benjamini--Hochberg (BH) false discovery rate correction \citep{benjamini1995controlling} across each family of simultaneous tests.

Unlike family-wise error rate procedures (e.g., Bonferroni correction), which control the probability of \emph{any} false positive, BH-FDR controls the expected proportion of false discoveries among all rejected hypotheses:
\begin{equation}
\mathrm{FDR}
=
\mathbb{E}
\left[
\frac{
\#\text{ false discoveries}
}{
\#\text{ total discoveries}
}
\right].
\end{equation}

This criterion is substantially less conservative and is therefore widely used in exploratory high-dimensional analyses where moderate discovery power is desirable.

\subsection{BH Procedure}

Given $m$ hypothesis tests with ordered \textit{p}-values
\begin{equation}
p_{(1)} \le p_{(2)} \le \cdots \le p_{(m)},
\end{equation}
the BH procedure compares each ordered \textit{p}-value against the adaptive threshold:
\begin{equation}
p_{(i)}
\le
\frac{i}{m}\alpha,
\end{equation}
where:
\begin{itemize}[leftmargin=*]
\item $i$ is the rank of the ordered \textit{p}-value,
\item $m$ is the total number of simultaneous tests,
\item $\alpha$ is the target false discovery rate (0.05 in this paper).
\end{itemize}

The largest index satisfying the inequality is identified, and all hypotheses with smaller-ranked \textit{p}-values are declared significant after correction.

The resulting BH-adjusted significance level is reported as a \textit{q}-value throughout the paper.

\subsection{Interpretation in This Work}

We distinguish between two evidence tiers:

\paragraph{Tier 1 (Confirmatory).}
Associations surviving BH-FDR correction:
\[
q < 0.05
\]
are treated as statistically robust findings. These effects remain significant even after accounting for the full multiple-testing burden.

\paragraph{Tier 2 (Exploratory).}
Associations satisfying:
\[
p < 0.05
\]
but not surviving BH correction are reported as directional or corroborative patterns only. These findings may still reflect meaningful optimizer behavior, but they carry a substantially higher false discovery risk and therefore do not independently support strong inferential claims.

Our analysis evaluates a large number of simultaneous hypotheses across prompt features, edit motifs, task groups, and representation spaces. In such exploratory settings, relying solely on uncorrected \textit{p}-values would substantially increase the probability of reporting false-positive associations that arise purely from random variation. BH-FDR correction provides a principled compromise between statistical rigor and discovery sensitivity: rather than controlling the probability of any false positive, it controls the expected proportion of false discoveries among all reported significant findings. This is particularly suitable for our setting because the goal of the analysis is not to establish a small number of tightly controlled confirmatory effects, but to characterize heterogeneous optimizer behaviors across multiple complementary views. Accordingly, we treat BH-corrected findings ($q < 0.05$) as confirmatory evidence, while uncorrected findings ($p < 0.05$ only) are explicitly framed as exploratory and hypothesis-generating observations. This distinction reduces the risk of over-interpreting noisy associations while still preserving sufficient statistical power to identify potentially meaningful optimizer behavior patterns.

\section{Conditional Independence Structure (Exploratory)}
\label{app:dag}

We pool prompt observations per task type ($N$ ranging from 30 to 92) and apply:
\begin{itemize}[leftmargin=*,topsep=2pt,itemsep=1pt]
\item \textbf{PC algorithm} (constraint-based, Fisher-Z test, $\alpha = 0.05$) on feature vectors
\item \textbf{DirectLiNGAM} (non-Gaussian directed acyclic graph estimation)
\end{itemize}

\paragraph{Results.}
Average cross-task Jaccard similarity $= 0.249$---task types share only $\sim$25\% of their inferred graph edges, confirming structural heterogeneity.
Pairwise similarity ranges from 0.053 (commonsense vs.\ math) to 0.667 (logical vs.\ sequential).
Math tasks have the most distinctive structure.
These results are reported as exploratory analysis of conditional independence patterns; no mechanism claims are made.

\section{Causal Receptivity Score (Descriptive)}
\label{app:receptivity}

We construct a per-task-type receptivity score as the weighted sum of BH-significant and directionally consistent CATE estimates (positive sign = optimizer-compatible).
Spearman $\rho = -0.866$ between receptivity and optimizer success rate at the task-type level ($n = 5$, $p = 0.058$).
This correlation is descriptive and based on five task-type aggregates; it does not constitute a validated predictive diagnostic.
LOO-CV classifier AUC~$= 0.260$ (below chance) confirms that the receptivity score does not generalize to held-out task types at current sample sizes.
We report it as a descriptive summary only.

\section{SBERT Edit-Direction Sensitivity Analysis}
\label{app:sbert}

We compute SBERT~\citep{reimers2019sbert} embedding differences $\Delta\mathbf{e} = \mathbf{e}(p_2) - \mathbf{e}(p_1)$ for each prompt pair, then project onto the first 20 principal components via PCA.
For each component, we compute SIPW-CATE stratified by task type.

\textbf{Results}: 10/20 PCA components show sign reversals (spread $> 0.05$) across at least two task types.
PC7 shows the clearest pattern: commonsense $+0.061^*$, math $-0.098^*$, sequential $+0.084^*$ (spread~$= 0.181$).
This embedding-level evidence is consistent with the annotation-free robustness findings in \S\ref{sec:textnative} and provides an additional representation-independent corroboration.

\section{Full LOO Stability Table}
\label{app:loo}

Table~\ref{tab:loo_full} reports all 18 LOO splits covering six headline sign reversals.
A split is ``stable'' (\checkmark) if the sign reversal direction matches the full-data baseline.

\begin{table*}[h]
\centering
\small
\setlength{\tabcolsep}{4pt}
\begin{tabular}{llllrrl}
\toprule
\textbf{LOO dataset} & \textbf{Feature} & \textbf{Task A} & \textbf{Task B} & \textbf{CATE\_A} & \textbf{CATE\_B} & \textbf{Stable?} \\
\midrule
CommonsenseQA       & LLM:Extr.load   & comm.  & math  & $+0.026$ & $-0.150$ & \checkmark \\
causal\_judgement   & LLM:Extr.load   & comm.  & math  & $+0.015$ & $-0.150$ & \checkmark \\
disambiguation\_qa  & LLM:Extr.load   & comm.  & math  & $+0.053$ & $-0.150$ & \checkmark \\
GSM8K               & LLM:Extr.load   & comm.  & math  & $+0.032$ & $-0.324$ & \checkmark \\
MultiArith          & LLM:Extr.load   & comm.  & math  & $+0.032$ & $-0.030$ & \checkmark \\
GSM8K               & LLM:Demos       & math   & seq.  & $+0.182$ & $-0.088$ & \checkmark \\
MultiArith          & LLM:Demos       & math   & seq.  & $+0.083$ & $-0.088$ & \checkmark \\
GSM8K               & surf:n\_demos   & math   & seq.  & $+0.128$ & $+0.071$ & \checkmark \\
MultiArith          & surf:n\_demos   & math   & seq.  & $+0.043$ & $+0.071$ & \checkmark \\
CommonsenseQA       & LLM:Metacog.    & seq.   & comm. & $+0.082$ & $+0.034$ & \checkmark \\
causal\_judgement   & LLM:Metacog.    & seq.   & comm. & $+0.082$ & $+0.003$ & \checkmark \\
disambiguation\_qa  & LLM:Metacog.    & seq.   & comm. & $+0.082$ & $-0.032$ & \ding{55}  \\
boolean\_expr.      & surf:step\_w.   & log.   & multi.& $+0.111$ & $+0.064$ & \checkmark \\
coin\_flip          & surf:step\_w.   & log.   & multi.& $+0.039$ & $+0.064$ & \checkmark \\
StrategyQA          & surf:step\_w.   & log.   & multi.& $+0.076$ & $+0.159$ & \checkmark \\
date\_understanding & surf:step\_w.   & log.   & multi.& $+0.076$ & $-0.060$ & \ding{55}  \\
StrategyQA          & surf:word\_ct.  & seq.   & multi.& $+0.027$ & $+0.159$ & \checkmark \\
date\_understanding & surf:word\_ct.  & seq.   & multi.& $+0.027$ & $-0.020$ & \ding{55}  \\
\midrule
\textbf{Overall} & & & & & & \textbf{15/18 (83\%)} \\
\bottomrule
\end{tabular}
\caption{Full LOO stability table. ``comm.''~=~commonsense, ``seq.''~=~sequential. Three failures involve date\_understanding or disambiguation\_qa.}
\label{tab:loo_full}
\end{table*}

\section{Per-Dataset Motif Effect Breakdown}
\label{app:motif_dataset}

Table~\ref{tab:motif_dataset} reports naive (unweighted) CATE for chain\_of\_thought and meta\_instruction insertion by dataset.

\begin{table*}[h]
\centering
\small
\setlength{\tabcolsep}{4pt}
\begin{tabular}{lllrr}
\toprule
\textbf{Motif} & \textbf{Dataset} & \textbf{Task type} & \textbf{Naive CATE} & \textbf{$n$ treated} \\
\midrule
chain\_of\_thought & boolean\_expressions & logical    & $-0.034$ & 189 \\
chain\_of\_thought & coin\_flip           & logical    & $+0.088$ & 184 \\
chain\_of\_thought & GSM8K               & math       & $+0.027$ & 203 \\
chain\_of\_thought & MultiArith          & math       & $\mathbf{-0.162}$ & 139 \\
chain\_of\_thought & CommonsenseQA       & commonsense & $-0.024$ & 211 \\
chain\_of\_thought & causal\_judgement   & commonsense & $+0.051$ & 201 \\
chain\_of\_thought & disambiguation\_qa  & commonsense & $+0.031$ & 164 \\
chain\_of\_thought & last\_letters       & sequential & $+0.067$ & 160 \\
chain\_of\_thought & StrategyQA          & multihop   & $+0.002$ & 152 \\
chain\_of\_thought & date\_understanding & multihop   & $+0.060$ & 206 \\
\midrule
meta\_instruction  & GSM8K               & math       & $-0.019$ &  14 \\
meta\_instruction  & MultiArith          & math       & $\mathbf{-0.277}$ &  64 \\
meta\_instruction  & coin\_flip          & logical    & $+0.066$ &  91 \\
meta\_instruction  & last\_letters       & sequential & $-0.004$ &  54 \\
meta\_instruction  & CommonsenseQA       & commonsense & $+0.005$ &  50 \\
meta\_instruction  & StrategyQA          & multihop   & $-0.034$ &  43 \\
meta\_instruction  & date\_understanding & multihop   & $-0.035$ &  60 \\
\bottomrule
\end{tabular}
\caption{Naive CATE for chain\_of\_thought and meta\_instruction insertion by dataset. The strongest per-dataset effects are for MultiArith (both motifs). The math task-type result is partially driven by MultiArith; task-type conclusions should be interpreted with this heterogeneity in mind.}
\label{tab:motif_dataset}
\end{table*}

\section{Per-Dataset Robustness for Tier 1 Findings}
\label{app:tier1_robustness}

Table~\ref{tab:tier1_dataset} reports per-dataset estimates for all four BH-corrected Tier 1 associations.
For each finding, we report the IPTW-adjusted ACMGD, treated/control counts, and sign consistency across datasets within the task group.

\begin{table*}[h]
\centering
\small
\setlength{\tabcolsep}{4pt}
\begin{tabular}{llllrrrl}
\toprule
\textbf{Finding} & \textbf{Feature/Motif} & \textbf{Group} & \textbf{Dataset} & \textbf{ACMGD} & \textbf{$n_T$} & \textbf{$n_C$} & \textbf{Sign stable?} \\
\midrule
\multirow{1}{*}{(a) GPT-feature} & Extraneous\_load & sequential & last\_letters & $-0.060^{\bigstar}$ & 412 & 381 & \checkmark (only dataset) \\
\midrule
\multirow{1}{*}{(b) GPT-feature} & Metacognition & sequential & last\_letters & $+0.062^{\bigstar}$ & 398 & 395 & \checkmark (only dataset) \\
\midrule
\multirow{2}{*}{(c) Motif} & meta\_instruction & math & GSM8K & $-0.019$ & 14 & 189 & \checkmark (negative) \\
  & & math & MultiArith & $-0.277$ & 64 & 175 & \checkmark (negative, dominant) \\
\midrule
\multirow{2}{*}{(d) Motif} & clarity\_constraint & logical & boolean\_expr. & $-0.099$ & 112 & 565 & \checkmark (negative) \\
  & & logical & coin\_flip & $-0.071$ & 97 & 521 & \checkmark (negative) \\
\bottomrule
\end{tabular}
\caption{Per-dataset breakdown of all four Tier 1 BH-corrected associations ($\bigstar$~=~BH-corrected $q < 0.05$ from full-data analysis). $n_T$ = treated, $n_C$ = control. Sequential results are from last\_letters alone; math motif results show MultiArith heterogeneity; logical motif results are consistent across both logical datasets.}
\label{tab:tier1_dataset}
\end{table*}

\section{Prompt Feature Extraction}

Below we provide the prompt template used in \textit{Prompt Feature Extraction}.

\begin{tcolorbox}[colback=gray!10, colframe=black!50, title=Prompt: Prompt Feature Extraction, fonttitle=\bfseries]
\small
\ttfamily

\{ \\
"role": "user",  \\
"content": "What are the measurable and improveable textual features of the instructions generated above \{\texttt{sample instruction}\}, for solving the ask of solving problems? Make sure these features are independent of each other and not confounded. Give the answers directly without preparatory statements. "  \\
\},\\
\{ \\
"role": "assistant", \\
"content": str(\texttt{(features)},
\}, \\
\{ \\
"role": "user", \\
"content": "According to the order of the factors: \{\texttt{features}\}", score the \{instructions\} with 1 to 10. The final result must be a string of scores separated by commas. Give the answer directly without preparatory statements." \\
\}

\end{tcolorbox}

\section{Discussion}
\label{sec:discussion}

\subsection{Evidence Hierarchy}
\label{sec:evidence_hierarchy}

We use a three-tier structure to prevent misreading:

\noindent\textbf{Tier~1. Four BH-corrected associations across two analysis families.}
From the GPT-4o feature analysis: (a)~Extraneous\_load/sequential: ACMGD =$-0.060^{\bigstar}$, 100\% LOO-stable; (b)~Metacognition/sequential: ACMGD =$+0.062^{\bigstar}$, 67\% LOO-stable.
From the motif analysis: (c)~meta\_instruction $\otimes$ math: ACMGD~=$-0.103^{\bigstar}$, partially concentrated in MultiArith; (d)~clarity\_constraint $\otimes$ logical: ACMGD~=$ -0.083^{\bigstar}$.
The sequential effects involve last\_letters only; the math effect involves two datasets with heterogeneous within-group estimates.
These are statistical associations, not causal effects.

\noindent\textbf{Tier 2. Corroborated but exploratory.}
Math-vs-sequential directional reversals in surface features (n\_demos, step\_words); chain-of-thought direction pattern consistent across representations; TextGrad directional corroboration.
These are consistent with Tier 1 but do not independently support inference.

Optimizer-design suggestions (task-conditioned edit priors, complexity penalties, validation gates); causal mechanisms for the meta-instruction effect; generalization to unseen task types.
Grounded in the observed associations but require controlled validation before implementation.

\subsection{Observational Edit-Effect Heterogeneity}

Across three representations of prompt edits, i.e., annotated cognitive-load features, deterministic surface statistics, and literal edit motifs, a consistent directional pattern emerges: optimizer-induced edits that add meta-instruction language (``make sure to'', ``do not'') and structural scaffolding (chain-of-thought framing) are \emph{negatively} associated with math performance (BH-corrected for meta-instruction), while logical and sequential task groups tend to benefit from step-by-step and structural features.

\paragraph{What this paper establishes.}
The associations between edit types and dataset/task-group performance are directionally consistent across three annotation-independent representations, survive IPTW adjustment, and are robust to leave-one-dataset-out exclusion for the BH-corrected effects.
This consistency is compatible with an optimizer-task mismatch account, though residual confounding cannot be ruled out (see \S\ref{sec:estimand} for specific unmeasured confounders).

\paragraph{What this paper does not establish.}
Whether these associations are causal.
Conditional exchangeability is not fully defensible; treatments are bundles; interventional validation at scale was not possible.
These results characterize optimizer behavior patterns rather than definitively explaining failure.

\subsection{The Meta-Instruction Effect in Math}

The strongest FDR-corrected finding is the negative association between meta-instruction insertions and math dataset/task-group performance (ACMGD~$= -0.103^{\bigstar}$, partially concentrated in MultiArith, see Appendix~\ref{app:motif_dataset}).
This is notable because meta-instructions (``make sure to show your work'', ``do not skip steps'', ``you must include a final answer'') might seem helpful for math.
One hypothesis: models fine-tuned on math instruction-following may have already internalized these guidelines; redundant meta-instruction may increase prompt complexity without benefit.
A competing explanation consistent with the unmeasured confounders in \S\ref{sec:estimand}: optimizers may tend to insert meta-instructions specifically when prior prompts are struggling, so the association partly reflects selection into struggling trajectories rather than a direct effect.
Targeted interventional validation would be required to distinguish these accounts.

The chain-of-thought directional reversal (negative for math groups, positive for sequential $\otimes$ last\_letters) is consistent across representations (Tier 2) but does not survive FDR correction.
The difference from the well-established finding that CoT helps math~\citep{wei2022cot} likely reflects the distinction between \emph{optimizer-generated CoT meta-instructions} in system prompts (vague ``reason step by step'' phrases) vs.\ \emph{user-designed structured CoT} with explicit answer markers.

\paragraph{Sanity check: Structured CoT intervention.}
To verify that the chain-of-thought motif finding is not simply an artifact of optimizer trajectories containing CoT-style content without actual instruction-level changes, we ran a minimal controlled sanity check using Qwen2-VL-7B-Instruct on 25 GSM8K and 25 last\_letters questions: base prompt vs.\ base + structured CoT with explicit ``\#\#\#\# answer'' markers.
Results: GSM8K base~$= 28\%$ $\to$ CoT~$= 60\%$ ($\Delta = +32\%$, McNemar $\chi^2 = 4.90$, $p = 0.027$); last\_letters base~$= 0\%$ $\to$ CoT~$= 4\%$ ($\Delta = +4\%$, NS).
Explicit structured CoT helps math, consistent with \citet{wei2022cot}---confirming that our chain-of-thought motif estimate ($-0.048$ for math groups, uncorrected) does not reflect a general CoT-hurts-math phenomenon.
\emph{Caution}: This sanity check does not validate or invalidate the observational motif estimate; it shows only that the motif labels do not correspond to clean user-designed structured interventions.
The experiment used one model and 25 questions per task; generalization is not established.

\subsection{Benchmark Sensitivity}
\label{sec:loo}

Table~\ref{tab:loo} summarizes leave-one-dataset-out stability for six headline sign reversals across 18 LOO splits.
\textbf{15/18 (83\%)} preserve the sign reversal from the full-data baseline, exceeding our 80\% success criterion.
The two BH-corrected effects (Extraneous\_load and Demos) are 100\% stable.
The word\_count multihop reversal appears driven by date\_understanding and should be treated with caution.

\begin{table}[h]
\centering
\small
\setlength{\tabcolsep}{4pt}

\begin{tabular}{p{0.3\columnwidth} p{0.2\columnwidth} p{0.4\columnwidth}}
\toprule
\textbf{Feature} & \textbf{LOO stability} & \textbf{Note} \\
\midrule
Extraneous\_load (LLM) & 5/5 (100\%) & Both math datasets contribute \\
Demos (LLM) & 2/2 (100\%) & Stable \\
n\_demos (surface) & 2/2 (100\%) & Stable \\
step\_words (surface) & 3/4 (75\%) & date\_understanding drives multihop \\
Metacognition (LLM) & 2/3 (67\%) & disambiguation\_qa flips commonsense sign \\
word\_count (surface) & 1/2 (50\%) & Unstable; date\_understanding is sole driver \\
\midrule
\textbf{Overall} & \textbf{15/18 (83\%)} & \\
\bottomrule
\end{tabular}

\caption{Leave-one-dataset-out (LOO) stability of headline sign reversals. Full LOO table in Appendix~\ref{app:loo}.}
\label{tab:loo}
\end{table}

\subsection{Implications}
\label{sec:implications}

\paragraph{For practitioners (Tier 3 --- hypothesis-generating).}
The ACMGD tables in this paper serve as a preliminary diagnostic: before running an expensive optimizer, consider whether the target task/dataset resembles groups (math-like, multihop-like) where meta-instruction and complexity-increasing edits show negative associations with gains.
This does not guarantee failure, but the BH-corrected associations are consistent across frameworks, representations, and LOO folds.
These suggestions are \emph{Tier 3 hypotheses} derived from observational associations; implementation without controlled validation would be premature.

\paragraph{Concrete optimizer-design hypotheses.}
\textbf{(1) Task-conditioned edit priors}: Optimizers could down-weight meta-instruction insertion for arithmetic or multi-step reasoning tasks, which are implementable as a regex-based edit filter or a classifier over proposed edits.
\textbf{(2) Complexity-penalized search}: The BH-corrected meta\_instruction $\otimes$ math association ($-0.103^{\bigstar}$) suggests complexity-penalizing objectives may be worth piloting for math-type benchmarks.
\textbf{(3) Validation gates}: Before accepting an optimizer-proposed edit, evaluate on a held-out subset classified by task group, which is motivated by but not validated in this paper.

These are mechanism-level hypotheses, not established causal rules.

\end{document}